# [Vital Videos](): A public dataset of videos with PPG and BP ground truths

Toye Pieter-Jan ([pjtoye@gmail.com]() or [requests@vitalvideos.org]())

Vital Videos project, **vitalvideos.org** for up to date information and new datasets

---

**d7d98ef704c040dca286161dbf53a9e3**

Gender: M

Age: 29

Fitzpatrick: 2

Blood Pressure: 128/74

Min HR: 74

Max HR: 97

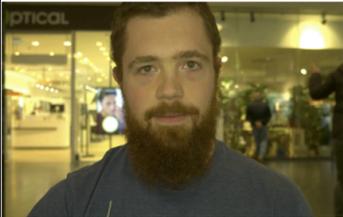 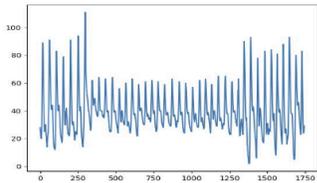 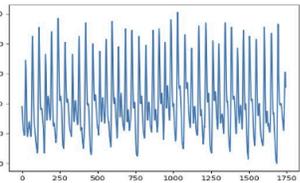

---

**acaefa9d52bd4dc486b8b4b781226e12**

Gender: M

Age: 26

Fitzpatrick: 2

Blood Pressure: 141/79

Min HR: 63

Max HR: 82

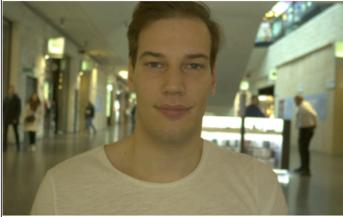 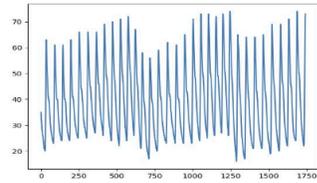 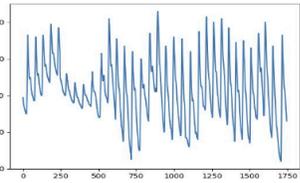

---

**d9e2735cddb1495bb2ccc8aff4d93f14**

Gender: F

Age: 28

Fitzpatrick: 2

Blood Pressure: 112/66

Min HR: 63

Max HR: 65

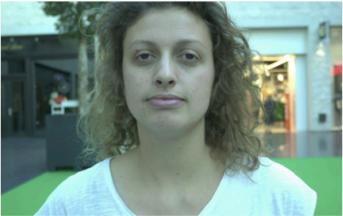 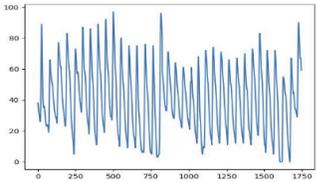 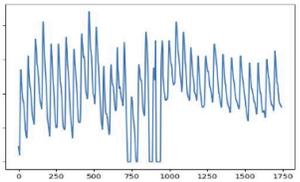

**Abstract**

We collected a large dataset consisting of 850 unique participants. For every participant we recorded two 30 second uncompressed videos, synchronized PPG waveforms and a single blood pressure measurement. Gender, age and skin color were also registered for every participant.

The dataset includes roughly equal numbers of males and females, as well as participants of all ages. While the skin color distribution could have been more balanced, the dataset contains individuals from every skin color. The data was collected in a diverse set of locations to ensure a wide variety of backgrounds and lighting conditions.

In an effort to assist in the research and development of remote vital sign measurement we are now opening up access to this dataset.

**Introduction**

The use of smartphone cameras for remote vital sign measurement has the potential to make a significant impact on a global scale. This technology is particularly promising because of the widespread availability of smartphones.

The existence of large and high-quality public datasets is essential for the rate of progress. Public datasets eliminate the need for researchers to collect their own data, which lowers the bar to entry. Furthermore, they enable us to replicate each other's work and facilitate benchmarking.

While there are already several public datasets available for researchers to use (McDuff, 2023), some of these datasets are relatively small and primarily consist of young, male participants with lighter skin types. Additionally, to the best of our knowledge, there are no public datasets that include blood pressure measurements.

In an effort to address these limitations and contribute to the research field, we have created a large dataset that aims to minimize participant bias and provide a representative sample of the population. Our dataset includes **both** PPG waveforms and blood pressure measurements. By providing researchers with access to this data, we hope to make a valuable contribution to the field.

**Data collection process**

Participants were informed on the nature and goal of the project before consenting to have their data used for the research, development and validation of remote vital sign monitoring technology.

After providing consent, participants were seated at a distance of 50 to 70cm in front of a camera on a tripod. The blood pressure cuff and pulseoximeter were attached to the participant. It was important to attach the pulseoximeter to one of the fingers of the arm without the cuff, otherwise the inflation of the cuff and subsequent blood flow restriction would have resulted in a signal loss for the pulseoximeter.

The participant was instructed to look straight ahead and keep the finger as stable as possible, preferably resting on the knee. When the PPG waveform stabilized, a 30 second video was recorded together with the PPG waveform. Heart rate and SpO2 value as calculated by the pulse oximeter were also captured.

After the first recording, another 30 second recording was started during which the blood pressure was also measured. PPG waveform, heart rate and SpO2 values were also captured during this recording. It is of vital importance to measure the blood pressure during the recording and not before or after because the measurement itself can have an impact on the blood pressure values.

After going through the whole process, the resulting measurements were discussed with the participant, they were thanked for their help and received a small reward.

**Hardware**

***Pulse oximeter***

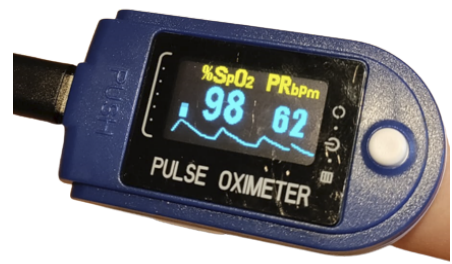

- Contec CMS50D+ a.k.a CMS50D-PLUS, not to be confused with CMS50D (without USB) or CMS50D-BT (uses Bluetooth)
- PPG sampling frequency: rated at 60Hz (*55-60*Hz in practice)
- HR/SpO$_2$ sampling frequency: 1Hz

***Blood pressure monitor***
- Omron M7
- Cuff on the upper arm (cuff was renewed after 500 participants)

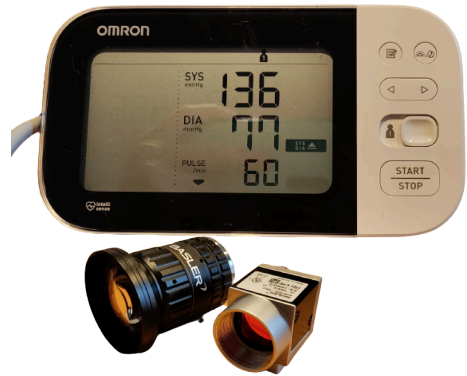

***Camera***
- Model: Basler acA1920-40uc
- Lens: Basler C11-1220-12M
- Connection: USB 3.0

The USB connection of this camera offered us the easiest path towards synchronization with the pulse oximeter PPG waveform. Its ability to deliver uncompressed frames was another key benefit.

***Camera settings***
- Resolution: 1920x1200
- FPS: 30
- Gain/ISO: 0 (because higher values can reduce video quality)
- Exposure time: 1/60 (16ms) which is half the frame rate, we don't need a shorter exposure time because there are no fast movements to be captured

***Video***
- Length: 30 seconds
- Codec: libx264 (lossless compression)
- Container: MP4
- File size: ~ 2GB / 30sec

**Software**
We created a custom Python tool with the following features:

- Live visual monitoring of camera and pulse-oximeter output
- Input of participant, location and scenario metadata
- Start and save recordings
- Assembling videos and ground-truth file

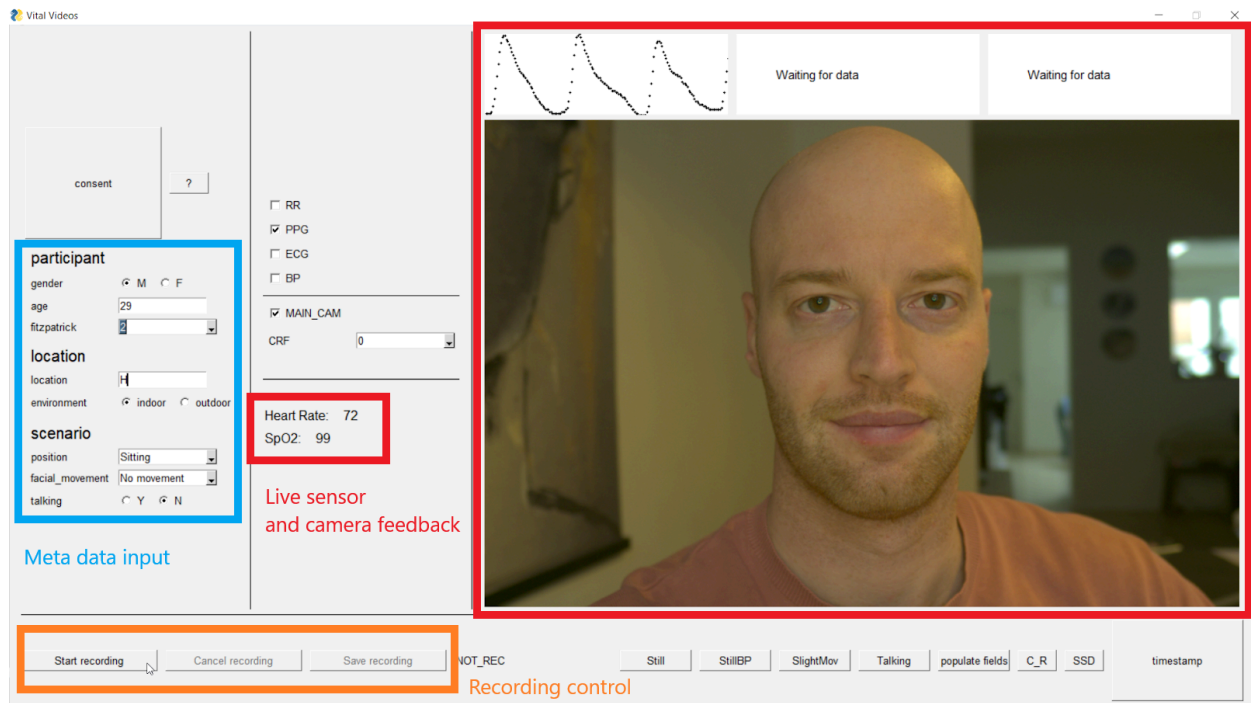

Screenshot of the user interface

**Locations & lighting**
Locations are very diverse and include a library, a retirement home, a shopping mall, a community center, … This resulted in recordings with artificial and natural light (behind or underneath large windows).

All locations were indoor because winter weather didn't allow for outdoor recording. Locations with labels KLL and KL don't have any natural light. When light intensity was too low a ring light was added to supply at least 150 lux to the face of the participant. Sometimes this resulted in rings that can be seen in the reflection of the participants' glasses.

Locations with labels KU, KUL, B, WZC and T have natural light sources. We always tried to seat the participant beneath or behind large windows to allow for as much natural light as possible. After a certain time (mostly around 17:00), the sun didn't provide at least 150 lux so in those scenarios we were required to add the ring light.

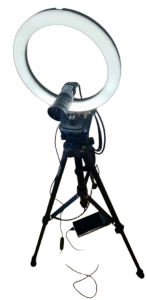

**Dataset descriptive statistics**

<u>VV-Small (100 participants, 390GB of raw video)</u>

This subset consists of 100 participants. These participants were handpicked from the complete dataset with the goal of maximal demographic (gender/age/skin color) and blood pressure diversity. Only the cleanest PPG ground truth waveforms were selected in this subset so that dataset users don't have to worry about correction for movement artifacts.

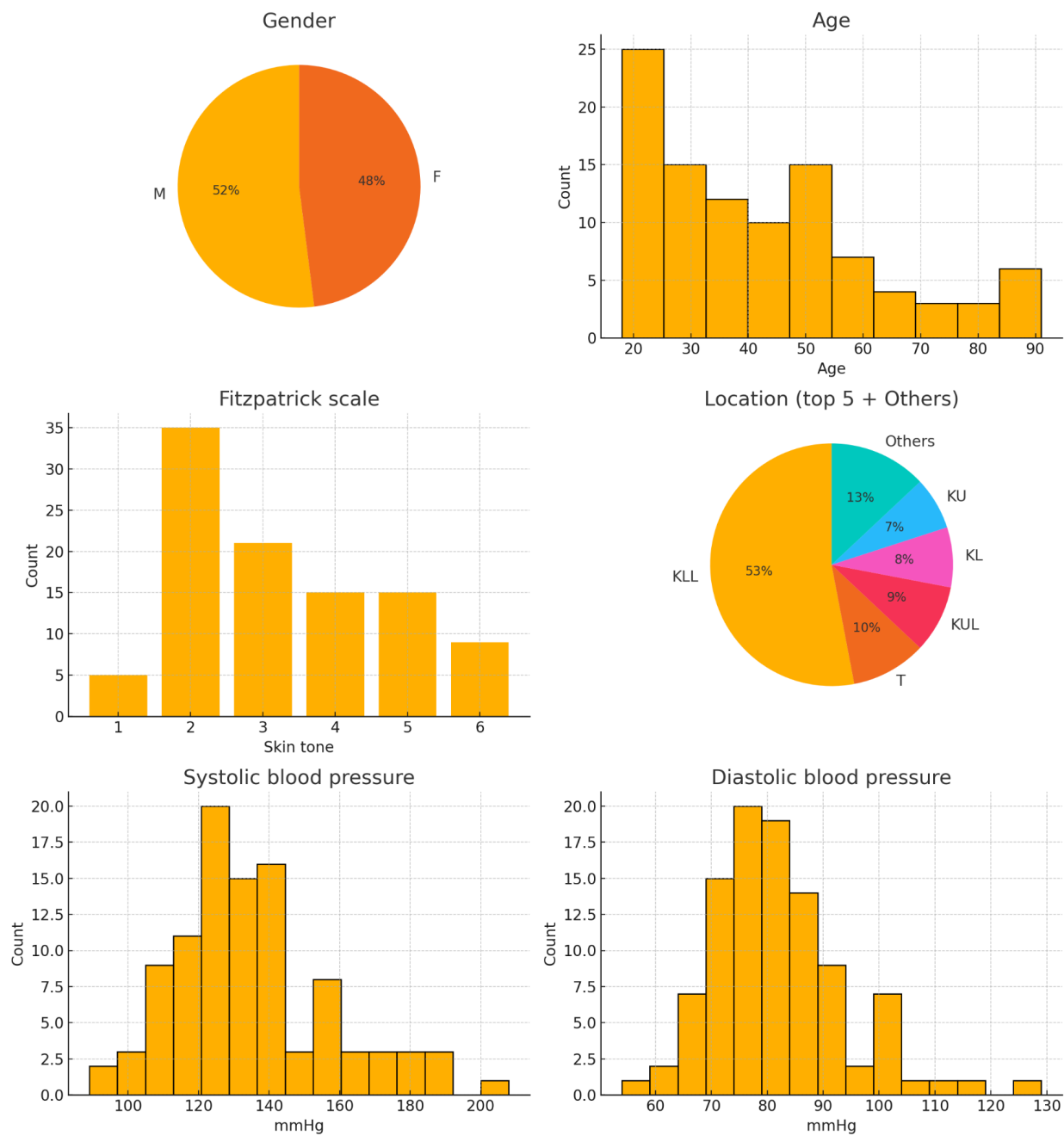

VV-Large (~ 850 participants, 3200GB raw video @ 500Mbps bitrate)
This is the complete dataset. Around 70-75% of all 60 second PPG waveforms are clean, 20% have 1 to 3 seconds of bad signal quality (caused by motion artifacts in the pulse oximeter) and around 5% of all PPG waveforms contain large disturbances (10-15 seconds).

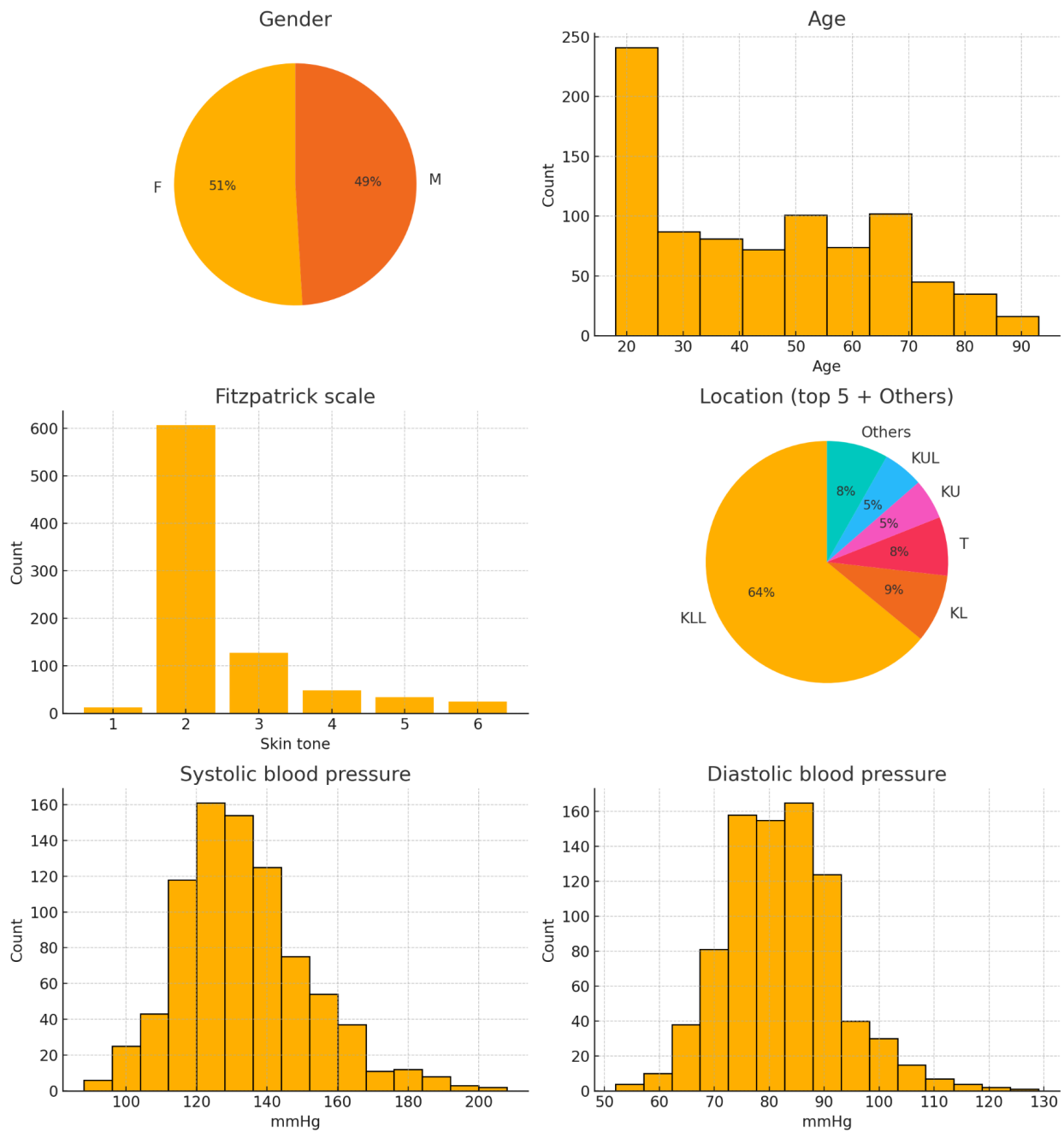

**Reflection and suggestions for future datasets**

**Cameras used**

Although the Basler camera has certain advantages (lossless recordings and easy synchronization), it would be better to use handheld smartphones because this will more closely align with future real world usage in terms of hardware used and natural camera movement.

The Basler camera also requires that you manually set focus which can result in lesser quality images if this is not done carefully or the participant moves forward/backward, this could also be resolved by smartphones who have auto-focus functionality.

Using smartphones for recording will bring its own issues, as they make synchronization with the PPG ground truth waveform a lot harder.

**Ground truths**

Capturing BP, PPG, HR and SpO$_2$ reference values is a good start, but more would have been better. ECG seems the most valuable addition, followed by GSR (galvanic skin response) and RR. Although much harder to collect, a blood drop or larger sample to determine glucose, cholesterol and other values would be extremely valuable to see what other parameters could possibly be extracted from a video of the face.

**Skin color labeling**

We labeled the skin color of the participants before each recording and were especially strict in labeling Fitzpatrick skin types 4, 5 and 6. Although this was done in a careful manner, it remains a subjective judgment. There is a better way of doing this by measuring skin color objectively with a colorimeter (Cortex Colorimeter DSM-4) but such devices are very expensive (~ 3.000 USD). Instead of labeling skin color before each recording, it would probably have been better to label the complete dataset in one go after recording everything, as this would probably have resulted in more consistent labels.

**Blood pressure measurements**

We measured blood pressure once for every participant during the recording. This could be improved upon by trying to elicit different blood pressure values (e.g. by calming or "stressing" the participant) and capturing these different values. Even just recording the blood pressure twice in short succession would be valuable to see if it was stable at the moment of recording.

**Recording time**

Two times thirty seconds is not a long recording time. Especially for some measurements like heart rate variability. For the participants however, one minute can feel quite long. To enable longer recordings without burdening the participant, we could entertain the participant using a short video.

**A larger variety of scenarios**

All participants were instructed to remain silent and look straight ahead. This makes it difficult to understand the impact of facial movement and talking on the accuracy of remote blood pressure measurements. It would be an improvement to include a larger variety of scenarios to study this.

**Demographics**

The complete dataset contains a nice 50/50 split of males and females. We are also quite satisfied with the age distribution. Skin color distribution is far from perfect, but given the recruiting region (Western-Europe) it is a decent result. To achieve a more balanced distribution of skin colors, it would be necessary to collect data on several continents.

**Size**

Another order of magnitude increase in size (upwards of 10.000 subjects) would be valuable to study the effect of scaling on the resulting accuracy of models but given our very limited resources, we had to limit its size.

**Use of pictures/frames**

You may **<u>not</u>** publish pictures of participants in papers or anywhere else. An exception is made for participant "3cf596e2bcc34862abc89bd2eca4a985". You are also allowed to use this picture in papers:

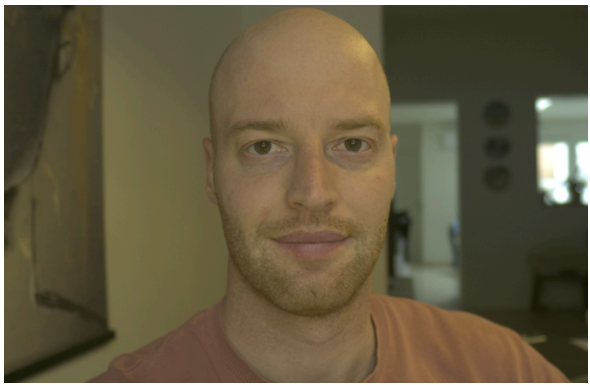

**Funding**

The creation of this dataset was funded from personal savings.

**Consent / IRB**
This study did not undergo Institutional Review Board (IRB) review due to our lack of affiliation with any university or organization, making the process both extra challenging and costly. All participants provided consent to participate in the dataset. This consent form grants permission to share the collected data with universities, research institutes, and companies for the purposes of research and product creation/validation.

**Availability**
Anyone can request to use the VV-Small  or VV-Large datasets for research or model development purposes.

If you or your organization doesn't directly or indirectly profit from the use of this dataset **and** you promise to publish your work publically (including code and models so that others can replicate your work) then you may apply to use a subselection of the dataset without any charge.

If you or your organization will directly or indirectly profit from the use of this dataset then an appropriate fee and license will be determined.

**Access procedure**
You can check out one sample here. The folder also contains some Python code to correctly process the .json ground truth files.

By following the instructions in this folder you can request a dataset.

**Publication**
If you use the Vital Videos dataset, please cite this publication:
Toye, P. (2023). Vital Videos: A dataset of face videos with PPG and BP ground truths.

**Conclusion**
We hope the dataset will enable a lot of new publications in the field thanks to the following characteristics:

- Its size which is at least an order of magnitude larger than most other public datasets
- Its balance in terms of gender and age categories
- Its inclusion of every skin color
- The diversity of locations and backgrounds
- Synchronized PPG waveforms

- Blood pressure ground truth
- Uncompressed video

And most of all: through its open nature which lowers the barrier to entry and enables broad benchmarking.

If you want to share your thoughts on the dataset or would like to discuss ideas for future datasets, please send an e-mail to pjtoye@gmail.com.

**Thanks**
We are grateful to all of our participants for their unique contributions to this dataset.

We would like to express our sincere thanks to Dr. Yannick Logghe who provided valuable insight into PPG waveforms and provided expert medical advice on many other topics. Dr. Philipp Roast provided many helpful comments and great suggestions on the creation of a quality RPPG dataset. We would also like to thank Brecht Dhuyvetters for providing inspiration and setting a good example. Gaëlle V. and Caroline V. were also most helpful by allowing access to two key locations to recruit participants.

We would like to thank both Peter H. Charlton and Daniel McDuff for providing comments to drafts of this paper.